\documentclass[runningheads]{llncs2}

\usepackage{amsmath} 
\usepackage{amssymb}  
\usepackage{graphicx} 
\usepackage{subcaption} 
\usepackage{listings}
\usepackage{enumitem}
\usepackage{hyperref} 
\usepackage{color} 
\usepackage{wrapfig}
\usepackage{tablefootnote}
\definecolor{light-gray}{gray}{0.4}

\DeclareMathOperator{\EX}{\mathbb{E}}

\title{Crawling in Rogue's dungeons\\ with (partitioned) A3C}

\author{Andrea Asperti, Daniele Cortesi, Francesco Sovrano}
\institute{
University of Bologna\\Department of Informatics: Science and Engineering (DISI)\\Mura Anteo Zamboni 7, 40127 Bologna, Italy.}

\begin{document}

\maketitle
\thispagestyle{empty}
\pagestyle{empty}

\begin{abstract}
Rogue is a famous dungeon-crawling video-game of the 80ies,
the ancestor of its gender.
Rogue-like games are known for the necessity to explore partially observable
and always different randomly-generated labyrinths, preventing any form of level replay.
As such, they serve as a very natural and challenging task for reinforcement learning, requiring the acquisition of complex, non-reactive behaviors involving memory and planning. In this article we show how, exploiting a version of
Asynchronous Advantage Actor-Critic (A3C)
partitioned on different situations, the agent is able to reach the stairs and descend to the next
level in 98\% of cases.
\end{abstract}

\section{INTRODUCTION}
In recent years, there has been a huge amount of work on the application of
deep learning techniques in combination with reinforcement learning (the so called 
{\em deep reinforcement learning}) for the development of automatic agents 
for different kind of games. Game-like environments provide realistic 
abstractions of real-life situations, creating new and innovative dimensions
in the kind of problems that can be addressed by means of neural networks. 
Since the seminal work by Mnih et al. \cite{Qlearning15} exploiting a combination of
Qlearning and neural networks (Deep Q-Networks, DQN)
in application to Atari games \cite{ALE13},
the field has rapidly evolved, offering several 
improvements such as Double Qlearning \cite{DoubleQlearning}
(correcting overestimations in the action value of the original version) 
to the recent breakthrough provided by the introduction of asynchronous methods, 
the so called A3C model \cite{A3C}. 

In this work, we apply a version of A3C to automatically move a player in the 
dungeons of the famous Rogue video game. Rogue was the ancestor of this gender
of games, and the first application exploiting a procedural, random creation of its
levels; we use it precisely in this way: as a generator of different kind of
labyrinths, with a reasonable level of complexity. Of course, the full game 
offers many other challenges, comprising collecting objects, evolving the rogue,
and fighting with monsters of increasing power, but, at least for the
moment, we are not addressing these aspects (although they may provide 
interesting cues for future developments). 

We largely based this work on the learning environment that was previously 
created to this aim in \cite{RogueinaboxA,RogueinaboxB}, and that allows a simple 
interaction with Rogue. At the same time, the extension
to A3C forced a major revision of the environment, that will be discussed in
Section~\ref{sec:rogueinabox-refactor}.

The reasons for addressing Rogue, apart from the fascination of this vintage 
game, have been extensively discussed in \cite{RogueinaboxA,RogueinaboxB} (see also \cite{DesktopDungeonsEnv}), and we just recall here the main motivations.
In particular, Rogue's dungeons are a classical example of Partially Observable Markov Decision Problem (POMDP), since each level is initially unknown and not entirely visible. 
Solving this kind of task is notoriously difficult and challenging \cite{SuttonBook}, since it requires an important amount of {\em exploration}.

The other important characteristic that differentiates it from other,
more modern, 3D dungeons-based games such as ViZDoom \cite{ViZDoom} or the Labyrinth in \cite{A3C} is precisely the graphical interface, that in the case of Rogue is ASCII-based. 
Our claim is that, at the current state of knowledge, decoupling 
vision from more intelligent activities such as planning can only 
be beneficial, allowing to focus the attention on the really challenging
aspects of the player behavior. 

\subsection{Achievements overview}\label{sec:overview}
Rogue is a complex game, where the player (the ``rogue'') is supposed to roam through many different levels of a dungeon 
trying to retrieve the amulet of Yendor. In his quest, the player must be able to:
1. explore the dungeon (partially visible, when you enter a new level);
2. defend himself from enemies, using the items scattered through the dungeon;
3. avoid traps;
4. avoid starvation, looking for and eating food inside the dungeon.

Currently, we are merely focusing on exploring the maze:
as explained in Section~\ref{sec:rogueinabox-refactor}
monsters and traps may be easily disabled in the game. 

\begin{wrapfigure}{R}{.4\textwidth}\label{fig:rogue}
\centering
\includegraphics[width=.4\textwidth]{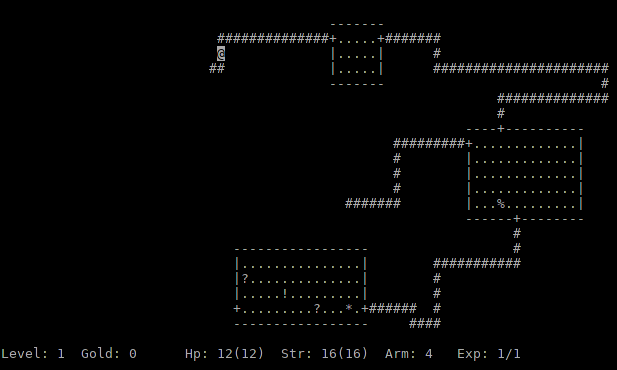}
\caption{The two dimensional ASCII-based interface of Rogue. }
\end{wrapfigure}

The dungeon consists of 26 floors (configurable) and each floor consists of up to 9 rooms of varying size and location, randomly connected 
through non linear corridors, and small mazes.
To reach a new floor the agent needs to find and to go down the stairs,
whose position is likely hidden from sight,
located in a yet unexplored room
and in a different spot at each new level.
Finding and taking the stairs are the main ingredients governing the agent movement: the only differences between the first floors and the subsequent ones 
are related to the frequency of meeting enemies, dark rooms, mazes or hidden doors.
As a consequence, we organized the training process on the base of a single level, terminating the episode as soon as the rogue takes the stairs.
In the rest of the work, when we talk about the {\em performance} of an agent, we refer to the probability that it correctly completes a {\em single}
level, finding and taking the stairs within a maximum of 500 moves\footnote{For a good agent, in average, little more than one hundred move are typically enough.}. The performance is measured on a set of 200 consecutive (i.e. random) games
and we show a comparison with previous work in table
\ref{tab:first-comparison}.
The results are not conclusive,
partly because the approaches rely on vastly different models.


\begin{table}[h!]
\vspace*{-1em}
\centering
\setlength{\tabcolsep}{0.5em} 
\begin{tabular}{|c|c|c|c|}\hline
agent       & random & DQN \cite{RogueinaboxA} & this work \\\hline
performance &   7\%\tablefootnote{The mobility resulting from brownian motion is always impressive.} & 23\%  & 98\% \\\hline
\end{tabular}
\vspace*{.5em}
\caption{}
\label{tab:first-comparison}
\vspace*{-3em}
\end{table}
There are essentially three ingredients behind this achievement:
\begin{enumerate}
\item the adoption of A3C as a base learning algorithm, in substitution of DQN; we shall diffusely talk about A3C in Section~\ref{sec:a3c}
\item an agent-centered, cropped representation of the state
\item a supervised partition of the problem in a predefined set of {\em situations}, each one delegated to a different A3C agent, sharing
nevertheless a common value function (i.e. a common evaluation of the state).\footnote{Source code and weights are publicly available at \cite{partitioned_a3c_repo}.}
We shall talk about situations in Section~\ref{sec:situations}
\end{enumerate}
While the adoption of A3C and the idea of experimenting with {\em situations} was a planned activity \cite{RogueinaboxB}, the shift to
an agent-centered view, as well as the choice of the agent situations have been mostly the result of trial-and-error, through an extremely long
and painful experimentation process.

\section{Related work}
As we mentioned in the introduction, there is a {\em huge} amount of research
around the application of deep reinforcement learning to video games . In this section we
shall merely mention some recent works that, in addition to those already mentioned,
have been a source of inspiration for our work, 
or the subject of different experimentations we performed. A few more works that seems to offer
promising developments \cite{ACER,GTN17} will be discussed in the conclusions. \\
Our current bot is essentially a partitioned multi-task agent in the sense of 
\cite{SunP99}. Its tree-like structure may be reminiscent of Hierarchical models \cite{Hierarchical16,HRL17,FeUdal17}, but they are in fact distinct notions. In Hierarchical models a Master cooperates with 
one or more Workers, by dictating them macro actions (e.g. ``reach the next room''), that are taken
by Workers as their objectives. The Master typically gets rewards from the environment  and  gives  
ad hoc, possibly {\em intrinsic}  bonuses  to Workers. The hope is to let top-level agents focus 
on planning while sub-parts of the hierarchy manage simple atomic actions, improving the learning process.\\
In our case, we simply split the task according to different situations the rogue may be faced with:
a room, a corridor, the proximity to stairs/walls, etc. (see Section~\ref{sec:situations} for details). 
We did several experiments with hierarchical structures, but so far none of them gave satisfactory 
result.

We also experimented with several forms of {\em intrinsic} rewards \cite{intrinsic04}, especially after passing to a rogue-centered view. Intrinsic motivations are stimuli received form the surrounding
environment different from explicit, extrinsic rewards, and that could be used by the agent for 
alternative form of training, learning to do a particular action because {\em inherently enjoyable}.
Examples are {\em empowerment} \cite{empowerment05} or {\em auxiliary tasks} \cite{AuxiliaryTasks16}.
In this case too, we have not been able to obtain interesting results.

\section{Reinforcement Learning Background}

\begin{wrapfigure}{R}{.5\textwidth}
\centering
\includegraphics[width=.5\textwidth]{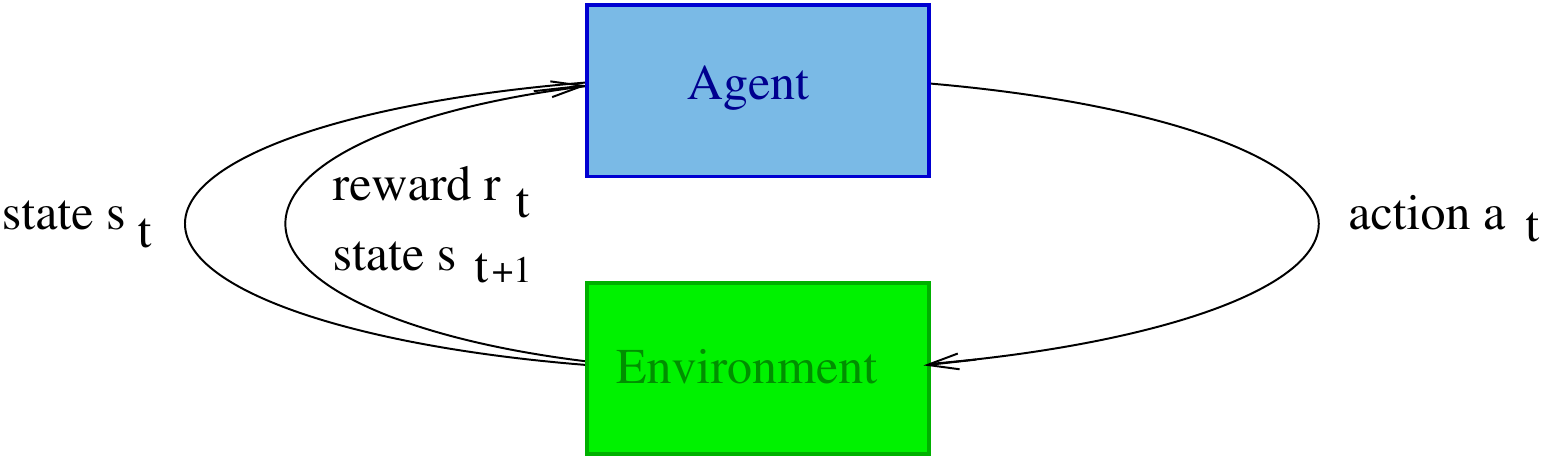}
\caption{Basic operations of a Markov Decision Process}
\label{fig:MDP}
\end{wrapfigure}

A Reinforcement  Learning  problem is usually formalized as
a Markov Decision Process (MDP).
In this setting, an agent interacts at discrete timesteps with an
external environment. At each time step $t$,  the
agent observes a state $s_t$ and choose an action $a_t$
according to some policy $\pi$, that is a mapping (or more generally
a probability distribution) from states to actions. As a result
of its action, the environment change to a new state $s'=s_{t+1}$; moreover
the agent obtains a reward $r_t$ (see Fig.~\ref{fig:MDP}).
The process is then iterated until a terminal state is reached.

The future cumulative reward
$R_t = \sum^\infty_{k=0}\gamma^{k}r_{t+k}$ is the total accumulated reward from
time starting at $t$. $\gamma \in \left[ 0, 1 \right]$ is the so called
{\em discount factor}: it represents the difference in importance between present and
future rewards. 

The goal of the agent is to maximize the expected return starting from an
initial state $s=s_t$. 

The  {\em action  value} $Q^\pi(s,a)  = \EX^\pi[R_t|s=s_t,a=a_t]$
is  the  expected return for selecting action $a$
in state $s_t$ and prosecuting with strategy $\pi$. 

Given a state $s$ and an action $a$, the optimal action value  function
$Q^*(s,a) = \max_\pi Q^\pi(s,a)$
is the best possible action value achievable by any policy.

Similarly, the {\em value} of state $s$ given a policy $\pi$ is
$V^\pi(s)  = \EX^\pi[R_t|s=s_t]$ 
and the optimal value function is
$V^*(s) = \max_\pi V^\pi(s)$.

\subsection{Q-learning and DQN}
The Q-function, similarly to the V-function can be represented by 
suitable function approximators, e.g. neural networks. We shall use the
notation $Q(s,a; \theta)$ to denote an approximate action-value function with parameters $\theta$.

In (one-step) Q-learning, we try to approximate the optimal
action value function: $Q^∗(s,a) \approx Q(s,a; \theta)$ by learning
the parameters via backpropagation according to a sequence
of loss function functions defined as follows:
\[L_i(\theta_i) = \mathbb{E}_{(s,a,r,s')\sim U(D)}\left[(r+\gamma \mathop{max}_{a'} Q(s',a',\theta_{i-1})-Q(s,a,\theta_i))^2 \right]\]
where $s'$ is the new state reached from $s$ taking action $a$
and $U(D)$ is the uniform distribution on stored transitions for
experience replay.

The previous loss function is motivated by the well know Bellman equation,
that must be satisfied by the optimal $Q^*$ function:
\[Q^*(s,a) = \mathbb{E}_{s'}[r_0 + \gamma max_{a'}Q^*(s',a')]\]
Indeed, if we know the optimal state-action values $Q^*(s',a')$
for next states, the optimal strategy is to take the action that maximizes 
$r_0 + \gamma max_{a'}Q^*(s',a')$.

Q-learning is an \textit{off-policy} reinforcement learning algorithm.
The main drawback of this method is that a reward
only directly affects the value of the state action pair
s,a that led to the reward. The  values  of  other  state  action  pairs  are
affected only indirectly through the updated value
$Q(s,a)$. The back propagation to relevant preceding
states and actions may require several updates,
slowing down the learning process.


\subsection{Actor-Critic and A3C}\label{sec:a3c}
In  contrast  to  value-based  methods,  policy-based
methods directly parameterize the policy
$\pi(a|s;\theta)$ and update the parameters $\theta$
by gradient ascent on $\EX[R_t]$.

The standard  REINFORCE  \cite{SuttonBook} algorithm
updates  the policy parameters
$\theta$ in the direction
$\nabla_\theta \EX [log \pi(a_t|s_t;\theta)R_t]$,
which is an unbiased estimate of $\nabla_\theta \EX[R_t]$.

It is possible to
reduce the variance of this estimate while keeping it unbiased
by subtracting a learned function of the state $b_t(s_t)$
known as a baseline. The gradient is then
$\nabla_\theta \EX [log \pi(a_t | s_t;\theta)(R_t - b_t)]$.

A learned estimate of the value function is commonly used
as the baseline $b_t(s_t) \approx V^\pi(s_t)$. In this
case, the quantity $R_t - b_t$ can be seen as
an  estimate  of  the {\em advantage}
of  action $a_t$ in state $s_t$ for policy $\pi$, defined as
$A^\pi(a_t|s_t) = Q^\pi(s_t, a_t) - V^\pi(s_t)$, 
just because $R_t$ is an estimate of $Q^\pi(s_t, a_t)$ and
$b_t$ is an estimate of $V^\pi(s_t)$. 

This approach can be viewed as an actor-critic architecture where the
policy $\pi$ is the actor and the baseline $b_t$
is the critic. 

A3C \cite{A3C} is a particular implementation of this technique based on the asynchronous 
interaction of several parallel couples of Actor and Critic. 
The experience of each agent is independent from that of the other agents, which stabilizes learning without the need for experience replay as in DQN.


\section{Neural Network Architecture}
Our implementation is essentially based on A3C. In this section we describe a novel technique that partitions the sample space into a predefined set of {\em situations}, each one addressed by a different A3C agent.
All of these agents contributes to build a common cumulative reward without sharing any other information, and for this reason they are said to be highly independent.
Each agent employs the same architecture, state representation and reward function. 
In this section we discuss: the situations (Sec.\ref{sec:situations}), 
the state representation (Sec.~\ref{sec:state-rep}), 
how we shaped the reward function (Sec.~\ref{sec:rewards}),
the neural network (Sec.~\ref{sec:nn}), hyper-parameters tuning (Sec.~\ref{sec:hyper}).

\subsection{Situations}
\label{sec:situations}
In our work, with the term \textit{situation} we mean the environment state used to discriminate which situational agent should perform the next action.
We experimented the four situations listed below, from higher to lower priority:
\begin{enumerate}
\item The rogue (the agent) stands on a corridor \label{situation_1}
\item The stairs are visible \label{situation_2}
\item The rogue is next to a wall \label{situation_3}
\item Any other case \label{situation_4}
\end{enumerate}
The situations are determined programmatically and are not learned.
This is a simplistic choice, mostly dictated by frustration:
in future work we plan to learn them in an end-to-end way.
When multiple conditions in the above list are met, the one with higher priority will be selected. For example, if the stairs are visible but the rogue is walking on a corridor, the situation is determined to be \ref{situation_1} rather than \ref{situation_2}, because the former has higher priority. \\
We define:
\begin{itemize}
\item {\em s4} as the configuration made of all the aforementioned situations
\item {\em s2} as the configuration made of situations \ref{situation_2} and \ref{situation_4}
\item {\em s1} as the configuration with no situations at all
\end{itemize}
We believe that situations may be seen as a way to simplify the overall problem, breaking it down into easier sub-problems.

\subsection{State representation}
\label{sec:state-rep}
The state is a $17 \times 17$ matrix corresponding to a cropped view of the map centered on the rogue (i.e. the rogue position is always on the center of the matrix). This representation has the advantage to be sufficiently small to be fed to dense layers (possibly after convolutions); moreover, it does not require to represent the rogue into the map.
In our experiments we adopted two variations of the above matrix.
The first (called \textit{c1}) has a single channel, resulting in a $17 \times 17 \times 1$ shape, and it is filled with the following values:
\begin{description}[style=multiline, leftmargin=1.5cm, labelindent=.8cm]
\item [4] for stairs
\item [8] for walls
\item [16] for doors and corridors
\item [0] everywhere else
\end{description}
The second (called \textit{c2}) is made of two channels (the stairs channel and the environment channel) and thus has shape $17 \times 17 \times 2$. The values used for \textit{c2} are the same of \textit{c1}.


\subsection{Reward shaping}\label{sec:rewards}
We designed the following reward function:
\begin{enumerate}
\item a positive reward ($+1$) is given when using a door never used before
\item a positive reward ($+1$) is given when, after an action, one or more new doors are found
\item a huge positive reward ($+10$) is given when descending the stairs
\item a small negative reward ($-0.01$) is given when taking an action that does not change the state (eg.: try to cross a wall)
\end{enumerate}
The chosen reward values are not random. In fact each floor contains at most $9$ rooms and each room has maximum $4$ doors, thus on each floor the cumulative reward of the rewards of type 1 and 2 can not exceed $9\cdot(4+2) = 54$. But what normally happens, in the episodes with the best return, is that only about $\frac{2}{3}$ of the cumulative reward is given by finding new rooms.
This is true because negative rewards are enough to teach the agent not to take useless actions and, in the meantime, they do not significantly affect the balance between room exploration and stair descent. \\
The result is that the agent is encouraged both to descend the stairs and to explore the floor, and this impacts positively and significantly on its performance.
In future work we plan to employ sparser reward functions that are not as
problem specific.

\subsection{Neural Network}
\label{sec:nn}


\begin{figure}[h!]
  \vspace*{-1.5em}
  \centering
  \includegraphics[width=0.8\linewidth]{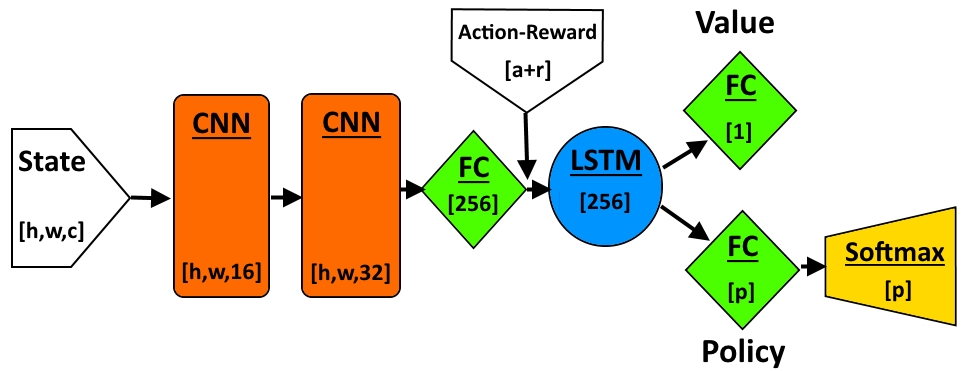}
  \caption{The neural networks architecture}
  \label{fig:arch}
  \vspace*{-1em}
\end{figure}
The neural network architecture we used is shown in figure \ref{fig:arch}. This network consists of two convolutional layers followed by a dense layer to process spatial dependencies and a LSTM layer to process temporal dependencies, and finally, value and policy output layers.
The convolutions have a ReLU activation, a $3 \times 3$ kernel with unitary stride and respectively 16 and 32 filters. Their output is flattened and fed to a FC with ReLU and 256 units. We call this structure: tower. \\
The tower input is the state representation described in Section~\ref{sec:state-rep} and its output is concatenated with a numerical ``one hot'' representation of the action taken in the previous state and the obtained reward. This concatenation is fed into an LSTM composed of 256 units. The idea of concatenating previous actions and rewards to the LSTM input comes from \cite{AuxiliaryTasks16}. \\
The output of the LSTM is then the input for the value and policy layers. \\
A network with the aforementioned structure implements an agent for each situation described in
Section~\ref{sec:situations}.
The loss is computed separately for each network, and corresponds to the A3C loss computed in \cite{AuxiliaryTasks16}.

\subsection{Hyper-Parameters Tuning}\label{sec:hyper}
Each episode lasts at most $500$ steps/actions, and it may end either achieving success (i.e. descending the stairs), or reaching the steps limit. Thus, the death state is impossible for the agent, since in our experiments monsters and traps have been disabled and $500$ steps are not enough to die for starvation. \\
Most of the remaining hyper-parameters values we adopted (for example the entropy $\beta = 0.001$) came from \cite{miyosuda_unreal}, an Open-Source implementation of \cite{AuxiliaryTasks16}, except the following:
\begin{description}[style=multiline, leftmargin=6cm, labelindent=.5cm]
\item [discount factor $\gamma$] 0.95
\item [batch size $t_{max}$] 60
\end{description}
We employed the same Tensorflow's RMSprop optimizer \cite{TF_RMSProp} available in \cite{miyosuda_unreal}, with parameters:
\begin{description}[style=multiline, leftmargin=6cm, labelindent=.5cm]
\item [decay] 0.99
\item [momentum] 0
\item [epsilon] 0.1
\item [clip norm] 40
\end{description}
The learning rate is annealed over time according to the following equation:
$\alpha = \eta \cdot \frac{T_{max} - T}{T_{max}}$, where $T_{max}$ is the maximum global step, and $T$ is the current global step. \\
The initial learning rate is approximatively $\eta = 0.0007$.



\section{Evaluation}
\begin{figure*}[h!]
  \centering
  \begin{subfigure}[t]{0.45\textwidth}
    \centering
    \includegraphics[width=\linewidth]{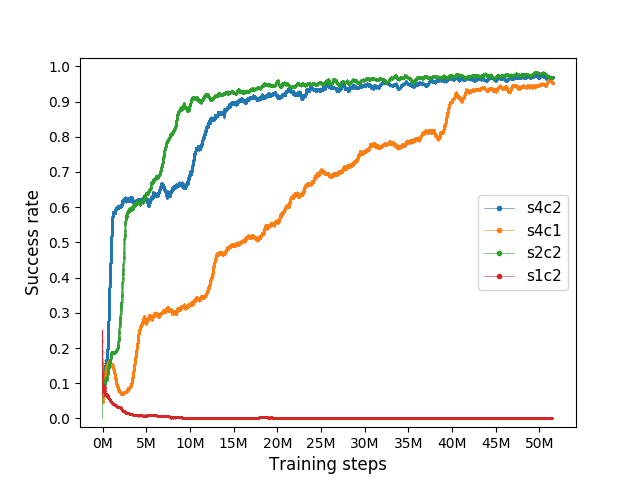}
    \caption{Success rate
  }
    \label{fig:learning-plot}
  \end{subfigure}~~%
  \hspace{.6cm}
  \begin{subfigure}[t]{0.45\textwidth}
    \centering
    \includegraphics[width=\linewidth]{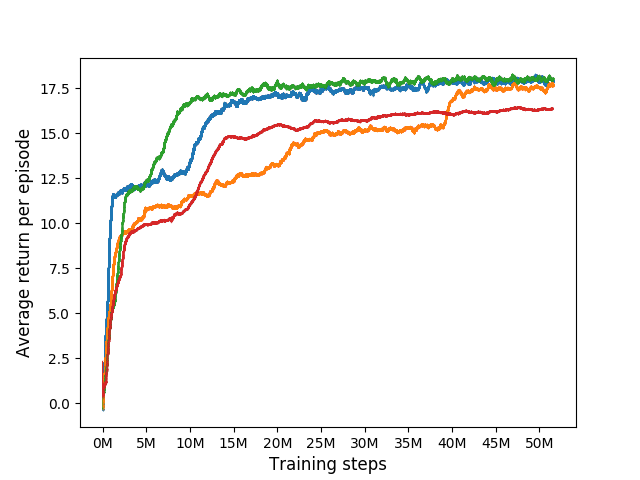}
    \caption{Avg. return per episode
    }
  \end{subfigure}
  \begin{subfigure}[t]{0.45\textwidth}
    \centering
    \includegraphics[width=\linewidth]{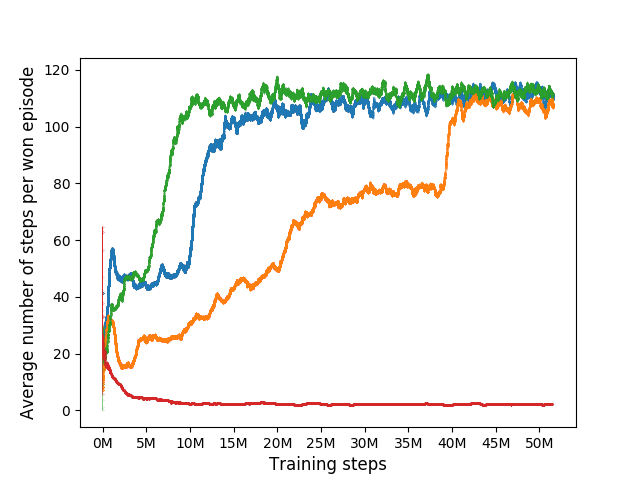}
    \caption{Avg. no. of steps per won episode
    }
  \end{subfigure}
  \caption{Results comparison. In the legend the labels {\em sX}
  denote the use of {\em X} situations,
  while {\em cY} a state representation with {\em Y} channels.
  Please see Sections \ref{sec:situations} and \ref{sec:state-rep} for details.}
  \label{fig:results}
\end{figure*}

For evaluation purposes we want to measure how often the agent is able to descend the stairs and to explore the floor. In our experiments, the final state is reached when the agent descend the stairs. For this reason, a good evaluation metric for a Rogue-like exploration-only system should be based at least on:
\begin{itemize}
\item the success rate: the percentage of episodes in which the final state is reached (an equivalent of the accuracy)
\item the number of new tiles found during the exploration process
\item the number of steps taken to win an episode
\end{itemize}
We evaluated our systems using an average of the aforementioned metrics over $200$ episodes.
The results we achieved are summarized in figure \ref{fig:results} and table \ref{tab:results}.\footnote{Source code and weights are publicly available at \cite{partitioned_a3c_repo}}

\begin{table}[h!b]
\setlength{\tabcolsep}{0.5em} 
\centering
\begin{tabular}{| l | l | l | l | l |}
\hline
Agent                             & \textit{s1-c2}  & \textit{s2-c2} & \textit{s4-c1}  & \textit{s4-c2} \\
\hline
Success rate                      &  0.03\%         & 98\%            &  96.5\%        & 97.6\% \\
\hline
Avg return                        &  16.16          & 17.97           &  17.66         & 17.99 \\
\hline
Avg number of seen tiles          &  655.02         & 386.46          &  365.88         & 389.27 \\
\hline
Avg number of steps to succeed  &  2.11           & 111.48          &  108.22        & 110.26 \\
\hline
\end{tabular}
\vspace{1em}
\caption{Learned policies evaluation.
With {\em sX} we denote the use of {\em X} situations
and with {\em cY} a state representation with {\em Y} channels.
Please see Sections \ref{sec:situations} and \ref{sec:state-rep} for details.}
\label{tab:results}
\end{table}

Our best agent\footnote{A video of our agent playing is available at
\url{https://youtu.be/1j6_165Q46w}}
shows remarkable skills in exploring the dungeon, searching for the stairs.

Using four situations instead of just two did not prove to be beneficial, however adopting a separate \textit{situation} (and hence a separate neural network) for the case when the stairs are visible was fundamental. In fact, as can be seen in Fig.~\ref{fig:results}, the policy learned by {\em s1-c2} completely ignored the stairs, thus achieving a very low success rate. \\
The experiment with 4 situations resulted in the development of the peculiar inclination for the agent of walking alongside walls. \\
Finally, state representation {\em c2} induced faster learning, but only
a slight increase in the resulting success rate.

\section{Refactoring the \textit{Rogue In A Box} library}
\label{sec:rogueinabox-refactor}
With this article, we release a new version \cite{partitioned_a3c_repo} of the \textit{Rogue In A Box} library \cite{RogueinaboxA,RogueinaboxB} that improves modularity, 
efficiency and usability with respect to the previous version. In particular, the old library was mainly centered around DQN-agents, that at the time looked as the most promising approach
for the application of deep reinforcement learning to this kind of games. With the advent 
of A3C and other techniques, we restructured the learning environment, 
neatly decoupling the interface with the game, supported by a suitable API,
from the design of the agents.\\
Other innovative features comprise:
\begin{enumerate}
	\item Screen parser and frames memory
    \item Communication between Rogue and the library
	\item Enabling or disabling monsters and traps
    \item Evaluation module
\end{enumerate}
Of particular note is the evaluation module, which provides statistics on the history of environment interactions, allowing to properly compare the policies of different agents.

\section{CONCLUSIONS}
In this article, we have shown how we can address the Partially Observable Markov Decision Problem behind the exploration of Rogue's dungeons, achieving a success rate of $98\%$, with a simple technique that partitions the sample space into situations.
Each situation is handled by a different A3C agent, all of them sharing a common
value function. The interest of Rogue is that the planar, ASCII-based, bi-dimensional interface permits to decouple vision from more intelligent activities such as planning:
in this way we may better investigate and understand the most challenging aspects of the player's behavior. 

The current version of the agent works very well, but still has some problems in cul-de-sac situations, where the agent should trace-back his path. Moreover, to completely solve the Rogue's exploration problem, \textit{dark rooms} and \textit{hidden doors} are also required to be handled. We predict that the main challenge is going to be provided by \textit{hidden doors}, since they are almost completely unpredictable and hard to detect even for a human. 
Different aspects of the game, such as collecting objects and fighting could also
be taken into account, possibly delegating them to ad-hoc situations.

In spite of the fact that the overall performance of our agent is really good, its design is not yet entirely satisfactory. In fact, too much intelligence about the game is built in, both in the design of situations, and especially in their identification and attribution to specific networks. Also the rogue-centered, cropped view introduces
a major simplification of the problem, completely by-passing the {\em attention}
problem (see e.g. \cite{Spatial15}) that, as discussed in \cite{RogueinaboxB}, was one of the interesting aspects of Rogue.


Currently, our efforts are going in the direction of designing an unsupervised version of the work described in this paper, where the agent is able to autonomously detect 
interesting situations, delegating them to specific subnets. As additional research topics, we are
\begin{itemize}
\item exploring the role of sample-efficiency in our context, along the lines of \cite{ACER}.
\item looking Multi-Task Adaptive Networks, following the ideas in \cite{GTN17}.
\end{itemize}

\bibliographystyle{splncs04}
\bibliography{biblio.bib,machine.bib}

\end{document}